\documentclass[journal]{IEEEtran}
\usepackage{amssymb, amsmath}
\usepackage{bm}
\usepackage{color}
\usepackage{graphicx}
\usepackage{cite}
\usepackage{booktabs}
\usepackage{multirow}
\usepackage{threeparttable}
\usepackage{enumerate}
\usepackage[english]{babel}
\usepackage{float}
\usepackage{placeins}
\usepackage{nomencl}
\usepackage{multicol}
\usepackage{commath}
\usepackage{makecell}
\usepackage{blindtext}
\usepackage{soul}
\usepackage{comment}
\usepackage{tablefootnote}
\usepackage[table]{xcolor}
\newcolumntype{C}[1]{>{\centering\arraybackslash}m{#1}}

\makenomenclature
\graphicspath{{Figures/}}

\usepackage{algorithmicx}

\usepackage[ruled, lined, linesnumbered, commentsnumbered, longend]{algorithm2e}

\makeatletter

\makenomenclature
\graphicspath{{Figures/}}

\begin{document}
\title{Initialization Is Critical: Advancing Federated Short-Term Load Forecasting under Load Heterogeneity via Model Initialization}
\author{Jianing Chen,~\IEEEmembership{Member,~IEEE}, 
        Vajiheh Farhadi,~\IEEEmembership{Member,~IEEE},
        Yan~Li,~\IEEEmembership{Senior Member,~IEEE},\\
        Thomas La Porta,~\IEEEmembership{Fellow,~IEEE} 
\thanks{J. Chen, Y. Li and T. F. La Porta are with the School of Electrical Engineering and Computer Science, The Pennsylvania State University, University Park, PA 16802, USA (e-mail: yql5925@psu.edu).}
\thanks{V. Farhadi is with the Department of Electrical and Computer Engineering, Bucknell University, Lewisburg, PA 17837, USA (e-mail: vf008@bucknell.edu).}}

\maketitle

\begin{abstract}

Short-term load forecasting (STLF) provides essential information for numerous applications in modern power systems. However, accurate STLF often relies on fine-grained smart-meter data from distributed users, raising increasing concerns about data privacy. Federated learning (FL) has therefore emerged as a promising privacy-preserving paradigm for STLF.
Nevertheless, this paper reveals structured heterogeneity in clients' load data. Specifically, clients exhibit different responses to exogenous factors and distinct temporal load profiles, which can degrade forecasting performance in FL. To mitigate these issues, this paper studies the role of model initialization in federated STLF, and proposes two initialization strategies from global and local perspectives.
For global model initialization, when auxiliary public load data are available, a pretrained initialization strategy is developed to initialize the global model before federated training, thereby reducing client drift during the training process. 
For local model initialization, we propose SLIAvg, a sequential local initialization strategy that promotes a more consistent training process by allowing participating clients to start from progressively adapted models within each communication round. Since the proposed strategies only modify the initialization process, they are compatible with most existing FL frameworks and privacy-enhancing techniques. Experiments on real smart-meter data with two representative forecasting architectures demonstrate that the proposed strategies effectively improve forecasting performance, as evidenced by reduced client drift, improved convergence behavior, and lower forecasting errors.
\end{abstract}

\section{Introduction}

\PARstart{A}{ccurate} short-term load forecasting (STLF) supports a wide range of critical applications in smart grids, spanning demand response, dynamic pricing, distribution-level state estimation, and economic dispatch \cite{gross_1987_shortterm, si2024robust, yang_2019_a}. With the rapid deployment of smart meters and advanced sensing infrastructure, STLF has increasingly shifted toward more granular scenarios, including residential-, building-, and user-level forecasting \cite{ mutanen_2011_customer}. However, the limited data from individual users make it difficult to train powerful single-user STLF models, whereas centralized training for a more capable model may raise privacy leakage concerns. This has motivated the use of federated learning (FL) for STLF \cite{zhang_2021_stochastic, luo2025privacy}.

FL collaboratively trains a global forecasting model through iterative local updates on distributed clients and server-side aggregation, without centralizing raw load data \cite{pmlr-v54-mcmahan17a}. This mechanism is effective when local data distributions are similar. Nevertheless, data heterogeneity across individual users remains a major challenge, often degrading the forecasting performance of FL \cite{sattler_2021_clustered}. In many conventional FL applications, data heterogeneity is commonly characterized by label-distribution skew, feature-distribution skew, or sample-size imbalance \cite{sun2023understanding, karimireddy2020scaffold}. However, load data heterogeneity among electricity users is manifested in more complex forms, driven by diverse behaviors, occupancy schedules, and weather sensitivities \cite{he2023privacy}. 
As a result, the local data distributions across users can differ substantially, driving local model updates away from the global optimization direction \cite{karimireddy2020scaffold}. This phenomenon, commonly referred to as client drift in FL, can slow down convergence and degrade the performance of predictive models \cite{yan2023rethinking, karimireddy2020scaffold}. Therefore, mitigating client drift is imperative for the practical deployment of FL in STLF. 

Existing studies have proposed effective solutions to improve the performance of federated STLF from different perspectives, including personalized learning schemes \cite{elgalhud2025federated, barja2025personalized, luo2025privacy}, client grouping and hierarchical framework \cite{fernandez2022privacy, he2023privacy, si2024robust}, robust aggregation strategies \cite{feng2025short} and enhanced forecasting model architectures \cite{liu2023fedforecast}. Personalized FL methods aim to address the heterogeneous consumption behaviors of different users by learning user-specific forecasting models. For instance, a personalized federated online learning framework is introduced in \cite{elgalhud2025federated} that allows distributed models to continuously learn from newly arriving data. 
A local personalization retraining step is proposed in \cite{barja2025personalized} to improve user-specific adaptation and enhance the overall performance of the FL model. 
In \cite{luo2025privacy}, personalized local models are constructed through tensor decomposition techniques.

Another important direction is client grouping and hierarchical FL frameworks \cite{tun2021federated, gholizadeh2022federated}. 
To mitigate client drift, clustering-based methods partition users based on load-pattern similarity, such that FL is performed within groups of users exhibiting similar consumption behaviors \cite{9632314, fernandez2022privacy}. Hierarchical FL is further improved in \cite{he2023privacy} by introducing privacy-preserving clustering at the intra-cluster level, thereby enhancing privacy protection within the hierarchical framework.
In addition, privacy-preserving clustering and anomaly detection mechanisms are introduced in \cite{si2024robust} to mitigate the effects of data incongruence and anomalies in local datasets.

Additionally, federated STLF has also been improved from other perspectives. 
In \cite{feng2025short}, an aggregation method that incorporates both global and local information is integrated into FL to improve STLF accuracy.
Enhanced forecasting architectures, such as iQGRU, are developed to better characterize the randomness of individual loads in probabilistic forecasting \cite{liu2023fedforecast}. 
A secure FL-STLF framework that combines differential privacy and gradient quantization is proposed in \cite{husnoo2024secure} to improve privacy protection and communication efficiency.

Recent studies in general FL have indicated that model performance is strongly influenced by model initialization \cite{chen2023pretrainingfl, nguyen2023initializationfl}. 
In centralized training, initialization is known to strongly affect convergence speed. 
However, its role in FL can be even more important. 
If the global model is randomly initialized, early-stage local updates are likely to be highly inconsistent across heterogeneous users. 
Such inconsistency can be amplified by multiple local training steps and then aggregated into unstable global updates, which can lead to severe client drift \cite{li2020convergence, karimireddy2020scaffold}. 
In contrast, an informative initialization can place the global model closer to a favorable parameter region, thereby promoting more stable federated optimization. 
This can make early local updates more aligned and improve the stability of aggregation. 
However, to the best of our knowledge, the role of model initialization has not been systematically investigated in federated STLF.

Building on this insight, this paper studies federated STLF by jointly considering data heterogeneity and model initialization. Unlike existing approaches that primarily improve forecasting architectures or rely on client clustering, the proposed framework mitigates client drift through model initialization at both the global and local levels. The main contributions are summarized as follows:
\begin{itemize}
    \item Load heterogeneity is systematically analyzed in terms of heterogeneous responses to exogenous factors and temporal load profiles, with quantitative indicators demonstrating substantial cross-user differences in both aspects.

    \item When auxiliary public load data are available, a pretrained global initialization strategy is proposed. The auxiliary data are used to pretrain the global model before FL, providing a more informative starting point that reduces client drift and improves forecasting accuracy.

    \item At the local level, a sequential local initialization strategy (SLIAvg) is proposed to mitigate client drift. By progressively adapting client initializations within each communication round, SLIAvg promotes smoother cross-client transitions and more consistent local updates.

    \item Comprehensive experiments are conducted on a real smart-meter dataset with two different forecasting architectures. 
    The results show that the proposed strategies effectively reduce client drift, improve convergence behavior, and achieve better performance across different client scales and model architectures.
\end{itemize}

It should be noted that the proposed methods are compatible with existing federated STLF methods, including advanced forecasting architectures, aggregation rules, personalization methods, since they only act on the initialization of the global model or local models. 
Additionally, the proposed initialization strategies can be further integrated with secure aggregation, differential privacy, or other privacy-enhancing techniques when stronger formal privacy guarantees are required \cite{bonawitz2017practical,zhu2019deep}.

The remainder of this paper is organized as follows. Section~II formulates the federated STLF problem and provides an overview of the main research components and their relationships. Section~III analyzes structured heterogeneity in load data and the resulting client drift. Section~IV presents two initialization strategies for federated STLF. Section~V presents numerical experiments, and Section~VI concludes the paper.
\section{Overview of the Federated STLF Framework}
\label{sec:overview}

Federated STLF enables multiple electricity users to collaboratively train a forecasting model without sharing their raw data. 
To formulate the federated STLF problem, consider a population of $K$ clients, where each client represents an electricity user participating in FL.
Each user $k$ locally stores a private dataset $\mathcal{D}_k$. 
For supervised STLF, each local model aims to predict future load based on historical observations and exogenous factors. 
Let $\mathbf{x}$ denote the input features, including historical load $\mathbf{y}^{\mathrm{hist}}$ and exogenous factors $\mathbf{u}^{\mathrm{exog}}$, and let $\mathbf{y}^{\mathrm{fut}}$ denote the target future load. 
Due to data availability, the exogenous factors considered in this paper include weather-related factors, such as temperature and humidity.

Let $f(\mathbf{w};\mathbf{x})$ denote a parametric forecasting model with parameter $\mathbf{w}$; representative architectures include Long Short-Term Memory (LSTM), Transformer, and Gated Recurrent Unit (GRU) \cite{hochreiter1997long,vaswani2017attention,cho2014learning}. The FL objective is formulated as
\begin{equation}
\begin{aligned}
\min_{\mathbf{w}} \quad
& F(\mathbf{w}) \triangleq \mathbb{E}_{k \sim P} \big[ F_k(\mathbf{w}) \big], \\
& F_k(\mathbf{w}) \triangleq \mathbb{E}_{(\mathbf{x}, \mathbf{y}^{\mathrm{fut}}) \sim \mathcal{D}_k}
\big[ \ell \left(f(\mathbf{w}; \mathbf{x}), \mathbf{y}^{\mathrm{fut}}\right) \big], \quad \forall k,
\end{aligned}
\end{equation}
where the random variable $k$ denotes a user index sampled from a user-population distribution $P$. 
The expectation $\mathbb{E}_{k \sim P}[\cdot]$ models how different users are weighted in the overall objective, e.g., according to local sample sizes or operational importance. 
The function $\ell (\cdot)$ denotes the loss function. 
In practice, $\mathbb{E}_{(\mathbf{x}, \mathbf{y}^{\mathrm{fut}}) \sim \mathcal{D}_k}[\cdot]$ is approximated by the empirical average over the local dataset $\mathcal{D}_k$, while $\mathbb{E}_{k \sim P}[\cdot]$ can be approximated through stochastic client sampling across communication rounds under partial participation.

\begin{figure}[t]
    \centering
    \includegraphics[width=6.0cm]
    {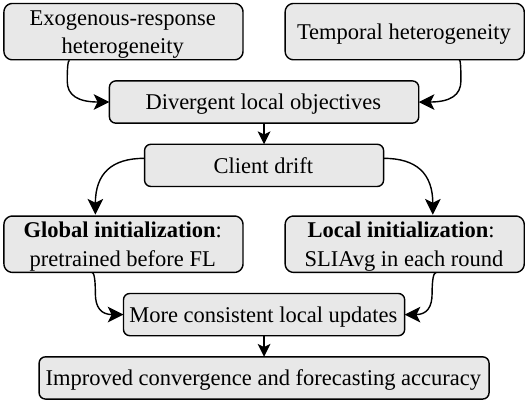}
    \caption{Overview of load heterogeneity analysis and the proposed initialization strategies.}
    \label{fig:framework_overview}
\end{figure}
Despite the privacy-preserving advantages of this framework, the performance of federated STLF can be significantly degraded by data heterogeneity. To investigate and address this challenge, this paper focuses on two main research components, as illustrated in Fig.~\ref{fig:framework_overview}. First, the sources and consequences of load data heterogeneity in federated STLF are analyzed. Specifically, load data across users exhibit two distinctive forms of heterogeneity: heterogeneous responses to exogenous factors and temporal heterogeneity. These forms of heterogeneity can lead to divergent local objectives and, consequently, client drift. Second, two model initialization strategies are developed at the global and local levels to mitigate the adverse effects of such heterogeneity. The proposed strategies promote more consistent local updates, thereby improving convergence and forecasting accuracy.

\section{Structured Load Heterogeneity and Client Drift in Federated STLF}

\subsection{Load Data Heterogeneity}

In standard FL literature, data heterogeneity is commonly characterized by feature-distribution skew and target-distribution skew\cite{yan2023rethinking, karimireddy2020scaffold}. 
Formally, it can be described as
\begin{equation}
p_k(\mathbf{x}, \mathbf{y}^{\mathrm{fut}}) \neq p(\mathbf{x}, \mathbf{y}^{\mathrm{fut}}),
\qquad k = 1,\dots,K,
\end{equation}
where the discrepancy often arises from differences in the marginal distributions, e.g.,
\begin{align}
\textbf{Feature-distribution skew:}\quad 
&p_k(\mathbf{x}) \neq p(\mathbf{x}) \label{fds}, \\
\textbf{Target-distribution skew:}\quad 
&p_k(\mathbf{y}^{\mathrm{fut}}) \neq p(\mathbf{y}^{\mathrm{fut}})\label{tds}.
\end{align}
Such feature-distribution skew or target-distribution skew can induce discrepancies among the local objective functions $F_k(\mathbf{w})$ across users.

In STLF tasks, non-IID client data arise from more complex mechanisms.
Beyond data heterogeneity \eqref{fds} and \eqref{tds}, load data heterogeneity often exhibits more structured and persistent patterns. Such heterogeneity can be characterized from two perspectives: \emph{1) \textbf{heterogeneous responses to exogenous factors}} and
\emph{2) \textbf{temporal heterogeneity}} across users.

\subsubsection{\textbf{Heterogeneous responses to exogenous factors}}
Each user $k$ observes a local load time series governed by
\begin{equation}
    \mathbf{y}^{\mathrm{fut}} = f_k^\star\big(\mathbf{y}^{\mathrm{hist}}, \mathbf{u}^{\mathrm{exog}}).
\end{equation}
The exogenous factors $\mathbf{u}^{\mathrm{exog}}$ play a critical role in shaping electricity demand.
Even when users may share a similar exogenous feature space, their true load-generation mechanisms $f_k^\star(\cdot)$ differ due to users' intrinsic characteristics. As a result, different users could exhibit heterogeneous responses to the same exogenous inputs.
This heterogeneity induces structured variation in the conditional distributions, i.e., 
\begin{align}\label{conditional_dist}
p_k(\mathbf{y}^{\mathrm{fut}} \mid \mathbf{u}^{\mathrm{exog}})
\neq
p_{k'}(\mathbf{y}^{\mathrm{fut}} \mid \mathbf{u}^{\mathrm{exog}}),
\quad k \neq k'.
\end{align}

In order to quantify the heterogeneous responses of users to exogenous factors, a linear response model is fitted to estimate the effects of exogenous variables on load demand for each user as follows,
\begin{align}\label{sen}
y_{k}^t
\approx
\alpha_k
+
\boldsymbol{\beta}_k^\top \mathbf{u}_t
+
\sum_{c \neq c_0}
\delta_{k,c}\mathbf{1}\{s_t=c\}
+
\boldsymbol{\gamma}_k^\top \mathbf{y}^{\mathrm{hist}}_{k,t},
\end{align}
where $y_{k}^{t}$ denotes the standardized load of user $k$ at time $t$, $\mathbf{u}_t$ denotes the continuous exogenous variables, $s_t$ denotes the categorical weather condition, and $\mathbf{y}^{\mathrm{hist}}_{k,t}$ contains historical load features. 
The category $c_0$ is treated as the reference weather condition. 
The parameters $\alpha_k$, $\boldsymbol{\beta}_k$, $\boldsymbol{\gamma}_k$, and $\delta_{k,c}$ are fitted using the local data of user $k$. 
Specifically, $\boldsymbol{\beta}_k$ and $\delta_{k,c}$ serve as empirical indicators of user $k$'s responses to exogenous conditions.
$\boldsymbol{\beta}_k$ characterizes the average sensitivity of the standardized load to continuous exogenous variables, while $\delta_{k,c}$ characterizes the average load shift associated with weather category $c$ relative to the reference category $c_0$.

\subsubsection{\textbf{Temporal heterogeneity}}
In addition to heterogeneous responses to exogenous factors, user load profiles
exhibit heterogeneous temporal structures.
Although the overall short-term periodicity of load dynamics is often shared across users,
the timing and evolution of these dynamics can differ due to individual  consumption behaviors.
Typical manifestations of temporal heterogeneity include daily load cycles with similar shapes but shifted peak hours, user-specific durations of load-state transitions, and stochastic timing variability in load responses across users.
Consequently, temporal heterogeneity can induce structured variation in the temporal conditional distributions, i.e.,
\begin{align}\label{temporal_dist}
p_k(\mathbf{y}^{\mathrm{fut}} \mid \mathbf{y}^{\mathrm{hist}})
\neq
p_{k'}(\mathbf{y}^{\mathrm{fut}} \mid \mathbf{y}^{\mathrm{hist}}),
\quad k \neq k'.
\end{align}

To characterize such temporal heterogeneity, the load profile of user $k$ can be described by a set of temporal load windows:
\begin{equation}
    \mathcal{Z}_k
    =
    \left\{
    \mathbf{z}_{k}^t
    =
    \left[
    y_{k}^{t-L+1},y_{k}^{t-L+2},\ldots,y_{k}^t
    \right]
    \right\}_{t=L}^{T_k},
\end{equation}
where $L$ denotes the window length and $\mathbf{z}_{k}^t$ represents the temporal evolution of user $k$ ending at time $t$.

To quantitatively measure temporal heterogeneity, the maximum mean discrepancy (MMD) is employed to compare the distributions of temporal load windows across users \cite{gretton2012kernel}. 
Let $Q_k$ denote the distribution induced by $\mathcal{Z}_k$. The temporal discrepancy between two users $k$ and $k'$ is measured by
\begin{equation}
    \mathrm{MMD}_{\kappa}
    \left(
    Q_k,
    Q_{k'}
    \right)
    \triangleq
    \left\|
    \mu_{Q_k}
    -
    \mu_{Q_{k'}}
    \right\|_{\mathcal{H}_{\kappa}},
\end{equation}
where $\mu_{Q}=\mathbb{E}_{z\sim Q}[\kappa(z,\cdot)]$ denotes the kernel mean embedding of distribution $Q$, and $\mathcal{H}_{\kappa}$ is the reproducing kernel Hilbert space induced by a positive definite kernel $\kappa(\cdot,\cdot)$. In this work, the Gaussian radial basis function kernel is adopted:
\begin{equation}
    \kappa(\mathbf{z},\mathbf{z}')
    =
    \exp
    \left(
    -
    \frac{
    \left\|\mathbf{z}-\mathbf{z}'\right\|_2^2
    }{
    2\sigma_{\kappa}^{2}
    }
    \right),
\end{equation}
where $\sigma_{\kappa}$ is the kernel bandwidth.

Given two finite window sets 
$\mathcal{Z}_k=\{\mathbf{z}_{k}^i\}_{i=1}^{n_k}$ and 
$\mathcal{Z}_{k'}=\{\mathbf{z}_{k'}^j\}_{j=1}^{n_{k'}}$, the empirical squared MMD is computed as
\begin{equation}
\begin{aligned}
    &\widehat{\mathrm{MMD}}_{\kappa}^{2}
    \left(
    \mathcal{Z}_k,\mathcal{Z}_{k'}
    \right)
    =
    \frac{1}{n_k^2}
    \sum_{i=1}^{n_k}
    \sum_{j=1}^{n_k}
    \kappa
    \left(
    \mathbf{z}_{k}^i,\mathbf{z}_{k}^j
    \right)
    \\
    &
    +
    \frac{1}{n_{k'}^2}
    \sum_{i=1}^{n_{k'}}
    \sum_{j=1}^{n_{k'}}
    \kappa
    \left(
    \mathbf{z}_{k'}^i,\mathbf{z}_{k'}^j
    \right)
    -
    \frac{2}{n_kn_{k'}}
    \sum_{i=1}^{n_k}
    \sum_{j=1}^{n_{k'}}
    \kappa
    \left(
    \mathbf{z}_{k}^i,\mathbf{z}_{k'}^j
    \right).
\end{aligned}
\end{equation}
A larger value of $\widehat{\mathrm{MMD}}_{\kappa}$ indicates a greater discrepancy between the temporal load patterns of two users.
Therefore, this metric provides a distribution-level measurement of temporal heterogeneity, thereby reflecting differences in peak timing, transition duration, and stochasticity across users.

In summary, heterogeneous responses to exogenous factors and temporal heterogeneity jointly give rise to structured non-IID behavior in STLF. 
Such non-IID behavior reflects user-specific predictive mappings $f_k^\star(\cdot)$ and can result in discrepancies among the local objectives $F_k(\mathbf{w})$ across clients. 
These discrepancies provide the basis for the client drift issue discussed in the next subsection.

\subsection{Client Drift in Federated STLF}

Client drift refers to the systematic deviation of local model updates from the global optimization direction, which makes it difficult to train an accurate predictive model. The structured non-IID behavior among users' data can exacerbate client drift. Specifically, when client drift is severe, persistent discrepancies arise between local and global gradients, i.e., $\nabla F_k(\mathbf{w}) - \nabla F(\mathbf{w})$,
indicating that local gradients may provide biased or inconsistent estimates of the global descent direction.
Thus, the aggregated update direction may no longer be well aligned with the descent direction of the global objective.
This mismatch undermines the performance of the final model.

Common FL strategies, such as FedAvg \cite{pmlr-v54-mcmahan17a} and its variants (e.g., FedAdam \cite{reddi2021adaptive}), are still insufficient to address this issue under heterogeneous data.
To clearly illustrate the mechanism of client drift, FedAvg is taken as a representative example due to its widespread adoption in FL.
The pseudocode of FedAvg is summarized in Algorithm~\ref{alg:fedavg}.

\begin{algorithm}
\caption{FedAvg}\label{alg:fedavg}
    \SetKwInOut{KwIn}{Input}
    \SetKwInOut{KwOut}{Output}

    \KwIn{Total clients $K$; participating clients $M$; local sample sizes $\{N_k\}_{k=1}^{K}$; learning rate $\eta$; communication rounds $T$; local steps $E$.}
    \KwOut{Global model $\mathbf{w}^{T}$.}

    \BlankLine
    Initialize global model $\mathbf{w}^{0}$\;
    \For{$t = 0,1,\dots,T-1$}{
    Server uniformly samples a subset of clients $\mathcal{S}_t \subseteq \{1,\dots,K\}$, where $\vert \mathcal{S}_t \vert = M$\;
    \For{$k \in \mathcal{S}_t$ \textbf{in parallel}}{
    $\mathbf{w}_k^{t,0} \leftarrow \mathbf{w}^{t}$\;
    \For{$e = 0,1,\dots,E-1$} {
    Compute gradient $\nabla F_k\!\left(\mathbf{w}_k^{t,e}\right)$ at $\mathbf{w}_k^{t,e}$\;
    $\mathbf{w}_k^{t,e+1} \leftarrow \mathbf{w}_k^{t,e} - \eta \nabla F_k\!\left(\mathbf{w}_k^{t,e}\right)$\;}}
    
    Server aggregates updates:
    \hspace{1em} $\mathbf{w}^{t+1} \leftarrow \sum_{k \in \mathcal{S}_t} \frac{N_k}{\sum_{j \in \mathcal{S}_t} N_j} \, \mathbf{w}_k^{t,E}$\;}
    \Return $\mathbf{w}^T$
\end{algorithm}

\begin{figure}
\begin{center}
\includegraphics[width=8.8cm]{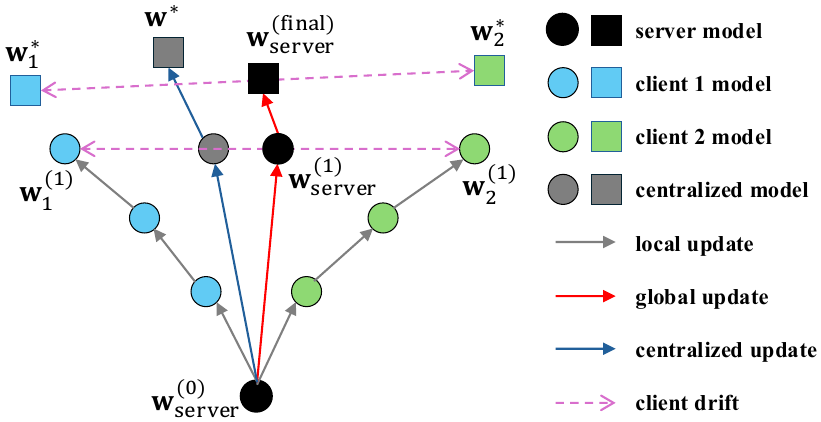}
\end{center}
\caption{Illustration of client drift in FedAvg compared with centralized optimization.
Two clients perform three local update steps ($K=2$, $E=3$).
Local updates (gray arrows) move toward client-specific optima $\mathbf{w}_1^*$ and $\mathbf{w}_2^*$ (blue and green squares).
Under heterogeneous local objectives, the aggregated server trajectory (red arrows) may deviate from the centralized optimization trajectory and move toward an averaged local solution $\mathbf{w}_{\text{server}}^{(\text{final})}$ (black square), rather than the centralized optimum $\mathbf{w}^*$ (gray square).}
\label{clientdrift}
\end{figure}

As shown in Algorithm \ref{alg:fedavg}, at each communication round, participating clients receive the current global
model parameters and perform multiple steps of local optimization on their
private datasets.
The server then aggregates the locally updated models, typically via a weighted average, to produce a new global model. While this procedure is effective under homogeneous data distributions, it exhibits pronounced client drift in the presence of data heterogeneity.
The mechanism of client drift is illustrated in Fig.~\ref{clientdrift} using a simplified example with two clients. Starting from the same initial server model, the two clients perform local updates toward different client-specific optima, $\mathbf{w}_1^*$ and $\mathbf{w}_2^*$, because their local objectives are heterogeneous. After aggregation, the server trajectory can therefore move toward an averaged local solution rather than following the centralized optimization trajectory toward the true centralized optimum $\mathbf{w}^*$. This illustrates how heterogeneous local updates can bias the aggregated FedAvg update and degrade the final global model.

\section{Advancing Federated STLF via Model Initialization under Load Heterogeneity}

To train an accurate predictive model within the FL framework, mitigating client drift is essential as discussed in Section II. Existing studies have demonstrated that model initialization plays a critical role in FL \cite{chen2023pretrainingfl, nguyen2023initializationfl}. A well-initialized model can accelerate convergence, improve training stability, and enhance generalization performance. Motivated by these facts, this paper proposes two strategies, aiming to alleviate client drift through both global and local model initialization: \emph{A) pretrained global initialization using auxiliary public load data}; and \emph{B) sequential initialization of local models}.

\subsection{Pretrained Initialization with Auxiliary Public Load Data}

A randomly initialized model tends to lie in a less informative region of the optimization landscape, where heterogeneous local data distributions can drive different clients toward substantially different descent directions, thereby leading to severe client drift. To quantify the client drift during FL, consider one communication round of FedAvg: 
\begin{equation}
    {\mathbf{w}}^{t+1}
    =
    \sum_{k=1}^{K} \rho_k
    \mathbf{w}_{k}^{t,E},
\end{equation}
where $\mathbf{w}_{k}^{t,E}$ denotes the local model of client $k$ after $E$ local steps, and $\rho_k$ is the aggregation weight of client $k$. The local-model divergence is defined as
\begin{equation}\label{cd}
    D_{\mathrm{loc}}^{t}
    \triangleq
    \sum_{k=1}^{K}
    \rho_k
    \left\|
    \mathbf{w}_{k}^{t,E}
    -
    {\mathbf{w}}^{t+1}
    \right\|_2^2.
\end{equation}
$D_{\mathrm{loc}}$ measures the divergence of local models around their weighted average after aggregation. Since all clients start from the same model at the beginning of each communication round, $D_{\mathrm{loc}}$ reflects the inconsistency of local model updates and can serve as an indicator of client drift \cite{li2020convergence}.

Although individual clients exhibit heterogeneous load behaviors, STLF tasks often share common temporal structures.
When auxiliary public load data are available, the model can be pretrained to capture dominant temporal load dynamics before federated training, thereby establishing a common representational prior across clients. 
Such pretrained initialization provides a more informative starting point for FL, allowing the model to start from a parameter region that is empirically closer to the final solution than random initialization. 
As a result, early-stage client updates become more consistent across heterogeneous clients, which helps reduce the value of $D_{\mathrm{loc}}$ and improve forecasting performance.

From the perspective of system operators or electric utilities, a limited amount of properly anonymized or publicly available load data may be accessible under privacy-preserving conditions. Such data may include historical public load records, anonymized measurements from representative users, simulation-generated load trajectories, or datasets voluntarily released by participating clients. Although these data are typically insufficient to train a fully optimized forecasting model, they can provide valuable global prior information for federated STLF.

Under this setting, a pretrained initialization strategy is proposed, in which the forecasting model is first trained in a centralized manner using available public or shared load data. The resulting pretrained model is then adopted as the initial global model $\mathbf{w}^{0}$ for subsequent federated training. Since the proposed strategy only modifies the initial global model parameters without introducing additional communication rounds or altering the client-side training protocol, it can be readily incorporated into existing federated optimization algorithms, such as FedAvg and FedAdam, and can be integrated with more advanced FL strategies. Moreover, because pretraining relies only on auxiliary public or shareable data, it does not require access to private client data during the pretraining stage, thereby avoiding additional privacy risks before federated training.

\subsection{Sequential Local Initialization without Pretraining Data}

In the standard FL setting, each participating client initializes its local model with the global model received from the server at the beginning of every communication round. 
Although this hard synchronization simplifies the training protocol, it may be suboptimal for STLF under strong data heterogeneity. 
Since all clients perform local training independently from the same initialization, their local models can quickly move toward different client-specific descent directions. 
After local training, these adapted models are directly averaged at the server. 
However, under strong data heterogeneity, local updates starting from the same initialization may point to substantially different or even conflicting directions, causing useful update components to be partially canceled and degrading global optimization \cite{li2023convergence}. 

\begin{figure}
\begin{center}
\includegraphics[width=8cm]{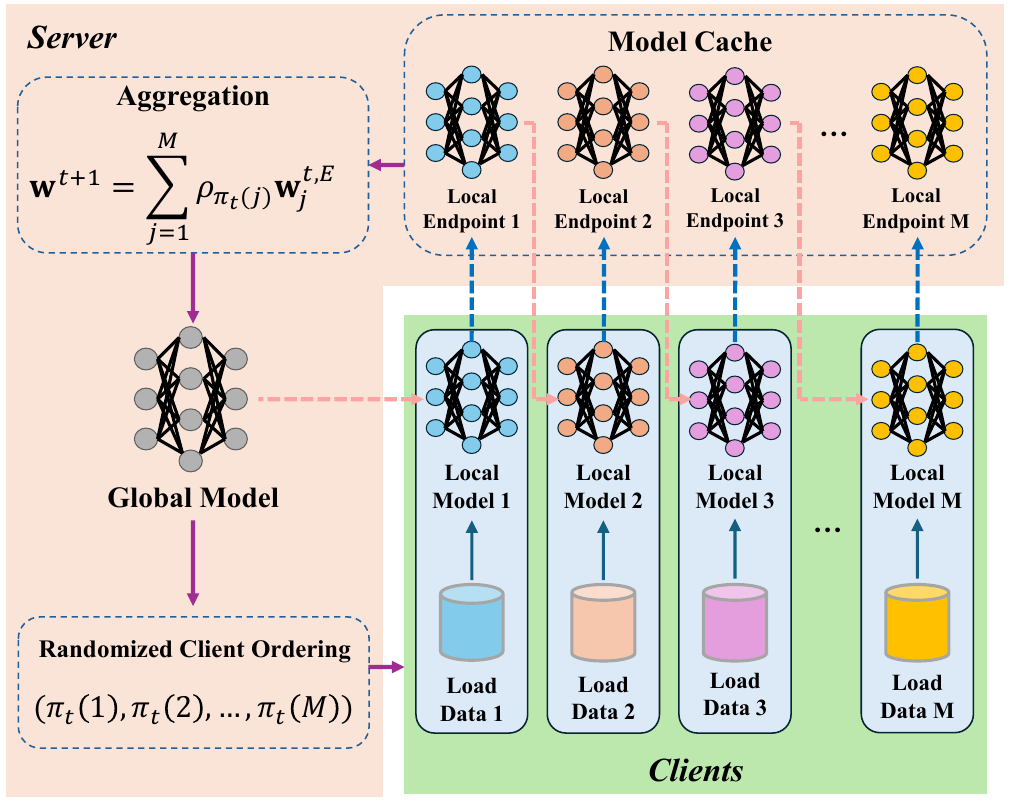}
\end{center}
\caption{Illustration of the proposed SLIAvg strategy with randomized client ordering, sequential local initialization, and server-side aggregation.}
\label{SLIAVG}
\end{figure}

To address this issue, a SLIAvg strategy is proposed to alleviate client drift, as illustrated in Fig.~\ref{SLIAVG}. 
Instead of broadcasting the same global model to all participating clients, SLIAvg uses the locally updated model of one client to initialize the next client through server coordination. Specifically, the server first sends the current global model to the first participating client in an order $\pi_t$. 
After local training, the updated model is returned to the server and then forwarded as the initialization for the next participating client. This process is repeated until all participating clients complete local training.

In this way, each client starts from a progressively adapted initialization that carries representations learned from preceding clients, allowing temporal features to be retained and refined before server-side aggregation. This progressive adaptation provides smoother transitions among local updates and reduces abrupt discrepancies caused by independent local training. 
As a result, local updates are expected to become more consistent across heterogeneous clients, which helps reduce client drift and improve forecasting accuracy.

In this regime, the initial model of each participating client is given by
\begin{equation}
    \mathbf{w}_{\pi_t(j)}^{t,0}
    =
    \begin{cases}
    \mathbf{w}^{t}, & j=1, \\[4pt]
    \mathbf{w}_{\pi_t(j-1)}^{t,E}, & j=2,\ldots,M.
    \end{cases}
\end{equation}
where $M$ is the number of participating clients. 

After all selected clients complete local training, their resulting local models are aggregated using the FedAvg rule. In the proposed SLIAvg, two issues are particularly important:

\paragraph{Randomized Client Ordering}

In SLIAvg, the initialization of each client depends on its position in the client sequence. 
Let the ordered participating clients in round $t$ be
\begin{equation}\label{order}
    \pi_t = \left(\pi_t(1), \pi_t(2), \ldots, \pi_t(M)\right),
\end{equation} 
Define the local update of the $j$-th participating client as
\begin{equation}
    \Delta_{\pi_t(j)}^{t}
    =
    \mathbf{w}_{\pi_t(j)}^{t,E}
    -
    \mathbf{w}_{\pi_t(j)}^{t,0}.
\end{equation}
According to the sequential initialization rule, the initial model of the $j$-th participating client is
\begin{equation}\label{seq_init}
    \mathbf{w}_{\pi_t(j)}^{t,0}
    =
    \mathbf{w}^{t}
    +
    \sum_{\ell=1}^{j-1}
    \Delta_{\pi_t(\ell)}^{t},
    \qquad j=1,\ldots,M.
\end{equation}
Thus, clients appearing later in the sequence are initialized with models that have accumulated more preceding local updates. 
If a fixed client order is repeatedly used, some clients may consistently occupy similar positions, leading to systematic order bias. 
To mitigate this issue, SLIAvg randomly shuffles the participating clients in each communication round.

\paragraph{FedAvg Aggregation for Avoiding Terminal-Endpoint Dominance}

Although randomized client ordering alleviates systematic order bias across rounds, the sequential initialization process remains order-dependent within each round. 
If the last local model were directly used as the next global model, as
$\mathbf{w}^{t+1}=\mathbf{w}_{\pi_t(M)}^{t,E}$,
the model may become increasingly adapted to later clients, and the representations learned from earlier clients may be gradually diluted or overwritten.

To avoid this issue, SLIAvg applies FedAvg aggregation to all sequentially trained local models. The global model is
\begin{equation}
    \mathbf{w}^{t+1}
    =
    \mathbf{w}^{t}
    +
    \sum_{\ell=1}^{M}
    a_{\ell}^{t}
    \Delta_{\pi_t(\ell)}^{t},
\end{equation}
where the position-dependent coefficient is
\begin{equation}
    a_{\ell}^{t}
    =
    \sum_{j=\ell}^{M}
    \rho_{\pi_t(j)}.
\end{equation}
Since $\rho_{\pi_t(j)}>0$ and $\sum_{j=1}^{M}\rho_{\pi_t(j)}=1$, 
\begin{equation}\label{weights}
    1=a_1^{t}
    >
    a_2^{t}
    >
    \cdots
    >
    a_M^{t}
    =
    \rho_{\pi_t(M)}.
\end{equation}
FedAvg aggregation prevents the global model from being solely determined by the terminal endpoint. By aggregating all intermediate local models, SLIAvg assigns relatively larger weights to earlier client updates as shown in \eqref{weights}, thereby mitigating excessive terminal-endpoint dominance.

The pseudocode of the proposed SLIAvg strategy is presented in Algorithm~\ref{init}.

\begin{algorithm}
\caption{SLIAvg}\label{init}
    \SetKwInOut{KwIn}{Input}
    \SetKwInOut{KwOut}{Output}

    \KwIn{Total clients $K$; participating clients $M$; learning rate $\eta$; communication rounds $T$; local steps $E$; local sample sizes $\{N_k\}_{k=1}^K$.}
    \KwOut{Global model $\mathbf{w}^{T}$}

    \BlankLine
    Initialize global model $\mathbf{w}^{0}$\;
    \For{$t = 0,1,\dots,T-1$}{
    Server uniformly samples a subset of clients $\mathcal{S}_t \subseteq \{1,\dots,K\}$, where $\vert \mathcal{S}_t \vert = M$\;
    Server draws a random permutation $\pi_t = \big(\pi_t(1),\pi_t(2),\dots,\pi_t(M)\big)$ of $\mathcal{S}_t$\;
    \For{$j = 1,2,\dots,M$ \textbf{in sequence}}{
    Initialize the local model $\mathbf{w}_{\pi_t(j)}^{t, 0}
    =
    \begin{cases}
    \mathbf{w}^{t} & j=1\\
    \mathbf{w}_{\pi_t(j-1)}^{t, E} & j>1
    \end{cases}$\;
    \For{$e = 0,1,\dots,E-1$} {
    Compute gradient $\nabla F_{\pi_t(j)}\!\left(\mathbf{w}_{\pi_t(j)}^{t,e}\right)$\;
    $\mathbf{w}_{\pi_t(j)}^{t, e+1} \leftarrow \mathbf{w}_{\pi_t(j)}^{t,e} - \eta \nabla F_{\pi_t(j)}\!\left(\mathbf{w}_{\pi_t(j)}^{t, e}\right)$\;}
    }
    Server aggregates all generated local endpoints: $\mathbf{w}^{t+1}\leftarrow\sum_{k \in\mathcal{S}_t}\frac{N_k}{\sum_{i \in\mathcal{S}_t} N_i}\mathbf{w}_{k}^{t,E}$\;}
    \Return $\mathbf{w}^T$
\end{algorithm}

\section{Numerical Examples}

In this section, numerical experiments are conducted to evaluate the effectiveness of the proposed strategies for STLF. The experiments are based on the public Low Carbon London dataset \cite{a2017_smart}, which contains smart-meter measurements collected by UK Power Networks together with detailed weather information. The smart-meter data are recorded at a half-hour resolution, resulting in 48 load observations per day. 

In this paper, 30 clients with load records from July 1, 2013, to February 16, 2014, are randomly selected from the dataset. Specifically, data from July 1, 2013, to January 15, 2014, are used for training, while data from January 16, 2014, to February 16, 2014, are used for testing. The data are processed using a sliding window of size 48, where the previous 48 half-hourly load observations are used to predict the current load. This setting follows the experimental setup in \cite{si2024robust}.

All experiments were implemented in Python 3.12 and PyTorch 2.11, and run on a workstation equipped with an Intel Xeon Gold 6542Y CPU at 2.90 GHz, 512 GB of host RAM, and two NVIDIA L40S GPUs, each with 48 GB of VRAM.

\subsection{Data Heterogeneity Analysis}

This section evaluates the heterogeneity of the load dataset adopted in this study.

\paragraph{Heterogeneous responses to exogenous factors} 

\begin{figure*}[!t]
\begin{center}
\includegraphics[width=15cm]{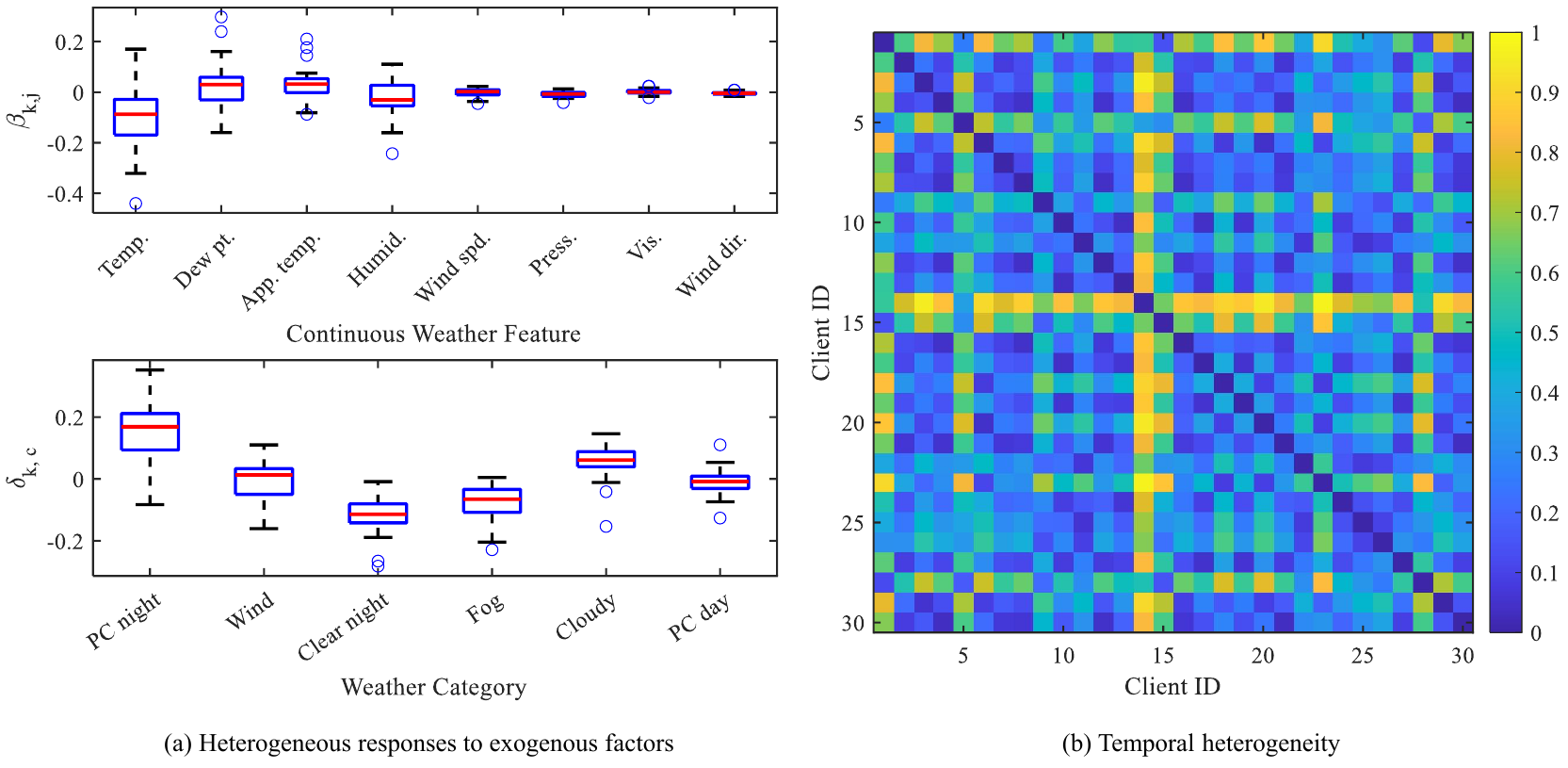}
\end{center}
\caption{Visualization of client heterogeneity in exogenous weather responses and temporal load-profile distributions. 
(a) Distributions of client-specific response coefficients to continuous and categorical weather factors. Temp., Dew pt., App. temp., Humid., Wind spd., Press., Vis., and Wind dir. denote temperature, dew point, apparent temperature, humidity, wind speed, pressure, visibility, and wind direction, respectively. 
PC day and PC night denote partly cloudy day and partly cloudy night. 
(b) Pairwise temporal heterogeneity matrix of client load profiles.}\label{hetero1}
\end{figure*}

Fig.~\ref{hetero1}(a) evaluates heterogeneous exogenous responses across clients based on \eqref{sen}. 
Since all clients share the same exogenous features at each time index, the estimated coefficients $\boldsymbol{\beta}_k$ and $\boldsymbol{\delta}_k$ are obtained under the same exogenous factors and are therefore comparable across clients. 

For continuous weather variables, the first box plot in Fig. \ref{hetero1}(a) presents several exogenous features which have clear cross-client variability, especially for temperature and dew point. For instance, $\beta_{k, j}$ related to temperature is negative, indicating that lower temperature is generally associated with higher electricity consumption, which is consistent with heating-driven demand in London. The spread of this coefficient also shows that the heterogeneity of this response varies substantially across clients. For categorical weather conditions, the second box plot in Fig. \ref{hetero1}(a) reports the client-specific response relative to the reference category, clear-day. The wide distribution of the coefficient in $\boldsymbol{\delta}_{k}$ reflects that different clients exhibit different load shifts under the same weather condition.

\paragraph{Temporal heterogeneity} Fig. \ref{hetero1}(b) presents the pairwise MMD matrix used to measure temporal heterogeneity across clients, where a larger value indicates stronger heterogeneity. The heatmap shows that the temporal distributions are not uniform across clients, as several client pairs exhibit large MMD values. This suggests that different clients have distinct temporal load behaviors. 

\subsection{Comparisons of Different Training Strategies}

To demonstrate the effectiveness of the proposed strategies, the following experimental settings are adopted:
\begin{itemize}
    \item To evaluate the generality of the proposed strategies, two representative forecasting architectures are considered, namely a Transformer model and a dual-channel LSTM model. For the Transformer model, $\mathbf{u}^{\mathrm{exog}}$ is fed into the encoder block, while $\mathbf{y}^{\mathrm{hist}}$ is used as the input to the decoder block. For the LSTM model, $\mathbf{u}^{\mathrm{exog}}$ and $\mathbf{y}^{\mathrm{hist}}$ are fed into two separate channels, respectively, and their learned representations are fused for final prediction.
    \item To ensure experimental reproducibility and improve the reliability of the comparison, the random seed is fixed to 42 for all steps involving weight initialization, data shuffling, and client sampling. This setting ensures that comparable runs use the same initial model weights and the same sampled clients participating in FL. By controlling these sources of randomness, the impact of stochastic variations on the experimental results is reduced.
    \item The reported evaluation metrics are mean squared error (MSE) and mean absolute error (MAE) \cite{feng2025short}. The reported results correspond to the best performance on the test dataset achieved during the entire training process.
\end{itemize}

\subsubsection{Evaluation of Pretrained Initialization with Auxiliary Public Load Data}

\begin{table}[htbp]
\caption{Effect of Pretraining Epochs on Test Loss ($K = 15$)}
\label{table_pretrain}
\centering
\renewcommand{\arraystretch}{1.05}
\setlength{\tabcolsep}{7pt}
\begin{tabular}{C{1.6cm} C{1.0cm} C{1.0cm} C{1.0cm} C{1.0cm}}
\toprule[1.5pt]
\multirow{2}{*}{Pretraining}
& \multicolumn{2}{c}{{Transformer}}
& \multicolumn{2}{c}{{LSTM}} \\ 
\cmidrule(lr){2-3} \cmidrule(lr){4-5}
Epoch & MSE & MAE
& MSE & MAE \\
\midrule
\multicolumn{5}{l}{\textbf{No pretraining}}\\
0
& 0.1363 & 0.1993
& 0.0897 & 0.1599 \\
\midrule
\multicolumn{5}{l}{\textbf{Pretraining without FL training data}} \\
1
& 0.1034 & 0.1708
& 0.0856 & 0.1567 \\
5
& \textbf{0.0928} & \textbf{0.1606}
& \textbf{0.0839} & \textbf{0.1566} \\
20
& 0.1073 & 0.1699
& 0.0873 & 0.1620 \\
\midrule
\multicolumn{5}{l}{\textbf{Pretraining with FL training data}} \\
1
& 0.0958 & 0.1724
& 0.0814 & 0.1536 \\
5
& 0.0980 & 0.1738
& 0.0779 & 0.1525 \\
20
& \textbf{0.0953} & \textbf{0.1658}
& \textbf{0.0753} & \textbf{0.1496} \\
\bottomrule[1.5pt]
\end{tabular}
\end{table}

Table~\ref{table_pretrain} reports the effect of pretrained initialization, where FedAvg is adopted. In this experiment, 15 clients are randomly selected to construct the training set, and 5 clients participate in each communication round. To examine the influence of pretraining data, two settings are considered: pretraining with data from 5 clients excluded from federated training, and pretraining with data from 5 clients included in federated training.

For both forecasting models under two settings, pretrained initialization improves the best MSE and MAE, reducing them from 0.1363/0.1993 to 0.0953/0.1658 for the Transformer model and from 0.0897/0.1599 to 0.0753/0.1496 for the LSTM model. However, when the pretraining dataset does not contain FL training data, increasing the number of pretraining epochs does not necessarily lead to better performance. For example, both models achieve their best performance at five pretraining epochs, whereas their performance degrades when the number of pretraining epochs is further increased to 20. This suggests that excessive pretraining on auxiliary data whose distribution differs from that of the FL training data may cause the model to learn client-specific patterns from the auxiliary clients. As a result, the pretrained model may move away from a favorable region for subsequent federated optimization, thereby reducing the effectiveness of pretrained initialization. 
This phenomenon further indicates that the effectiveness of pretrained initialization depends on whether the pretrained model provides a suitable starting point for subsequent federated optimization.

\begin{figure}
\begin{center}
\includegraphics[width=6.5cm]{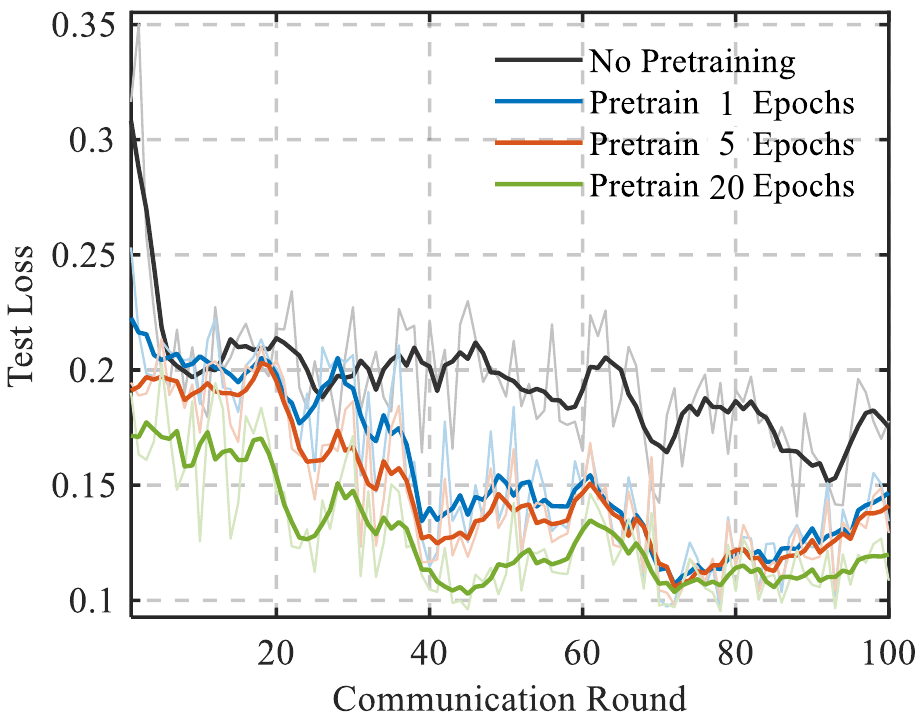}
\end{center}
\caption{The test loss of the Transformer-based models.}
\label{fig:loss_transformer}
\end{figure}

\begin{figure}
\begin{center}
\includegraphics[width=6.5cm]{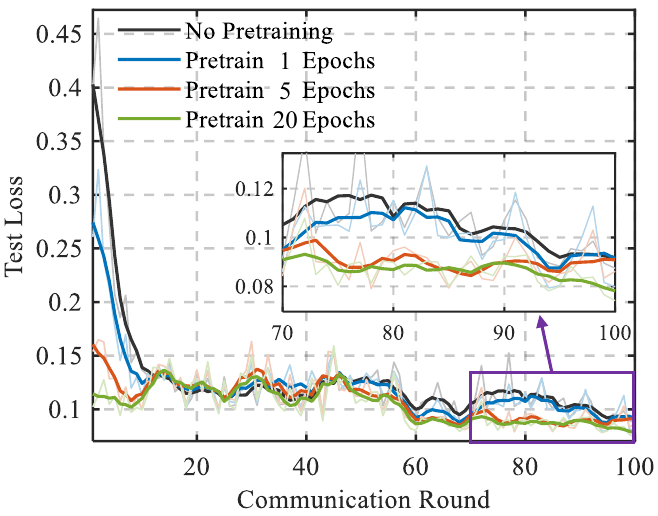}
\end{center}
\caption{The test loss of the LSTM-based models.}
\label{fig:loss_lstm}
\end{figure}

Figs.~\ref{fig:loss_transformer} and~\ref{fig:loss_lstm} illustrate the evolution of test loss over communication round for both forecasting models. Due to the presence of client drift, the original test loss curves exhibit considerable fluctuations during federated training. Therefore, to improve visual clarity, a moving-average smoothing with a window length of 5 is applied to the displayed curves. 

As shown in Figs.~\ref{fig:loss_transformer} and~\ref{fig:loss_lstm}, pretrained initialization leads to faster convergence in the early communication rounds, which is consistent with the behavior commonly observed in transfer learning. More importantly, pretrained initialization also achieves better forecasting performance, indicating that the auxiliary pretraining stage provides a better-aligned starting point for subsequent federated optimization. 

\begin{figure}
\begin{center}
\includegraphics[width=6.5cm]{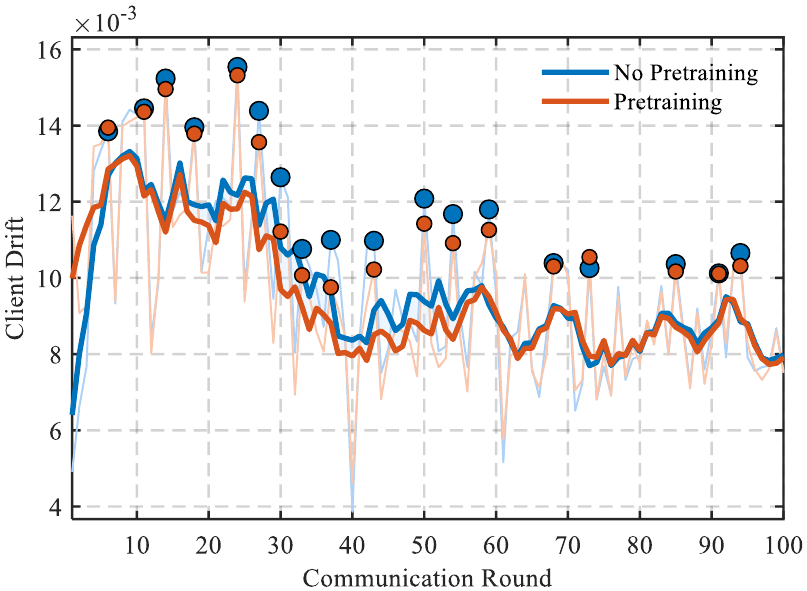}
\end{center}
\caption{Client drift of the Transformer model with and without pretraining.}
\label{fig:cd_transformer}
\end{figure}

\begin{figure}
\begin{center}
\includegraphics[width=6.5cm]{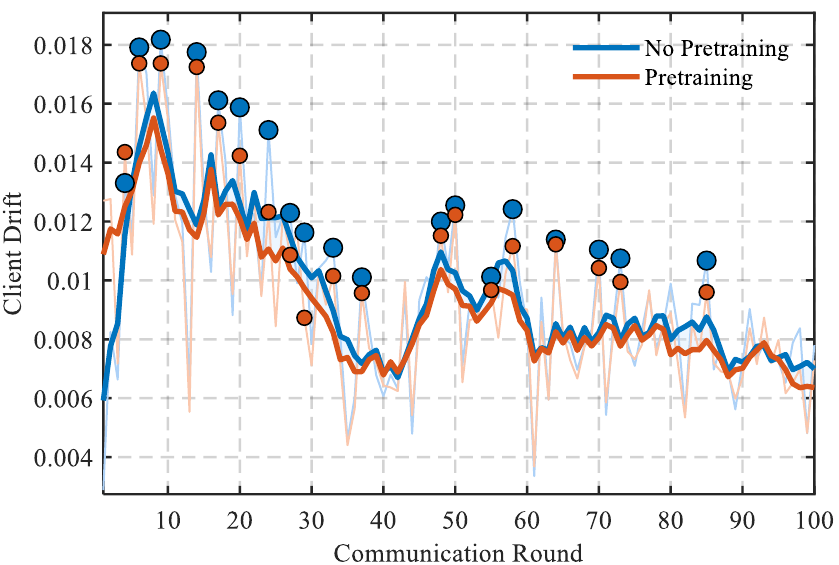}
\end{center}
\caption{Client drift of the LSTM model with and without pretraining.}
\label{fig:cd_lstm}
\end{figure}

To further explain the performance improvement, client drift is computed in Figs.~\ref{fig:cd_transformer} and \ref{fig:cd_lstm} via \eqref{cd}. A smaller drift indicates that local models move in more consistent optimization directions. Since the random seed is fixed, the same clients participate in each communication round under different settings. Therefore, the client-drift values are directly comparable across different settings. 

Compared with random initialization, pretrained initialization yields lower client drift in most subsequent rounds after the first few communication rounds, especially in the middle stage of federated training.
This suggests that pretrained initialization helps guide local optimization trajectories toward more consistent directions across clients, thereby improving the stability of server-side aggregation. 
The reduced client drift provides an important explanation for the lower test loss achieved by the proposed method.
It should be noted that the smaller client drift of the non-pretrained model in the first few communication rounds does not necessarily indicate better federated optimization. 
At this stage, the training loss remains large, and the model mainly learns coarse and common forecasting patterns from random initialization. 
As a result, local updates across clients may follow relatively similar descent directions, leading to smaller measured client drift. 
This phenomenon mainly reflects the early underfitted state of the model, rather than genuinely stable client-specific adaptation.

\subsubsection{Evaluation of Sequential Local Initialization}
\begin{table*}[htbp]
\caption{House-Level and Overall Forecasting Performance of Different Training Strategies}
\label{table2}
\centering
\renewcommand{\arraystretch}{1.3}
\begin{tabular}{C{1.3cm} C{1.3cm} C{1.3cm} C{1.3cm} C{1.3cm} C{1.3cm} C{1.3cm} C{1.3cm} C{1.3cm}}
\toprule[1.5pt]
\multirow{2}{*}{House ID} 
& \multicolumn{2}{c}{FedAvg} 
& \multicolumn{2}{c}{SLI} 
& \multicolumn{2}{c}{SLIAvg} 
& \multicolumn{2}{c}{\cellcolor{gray!20}Centralized} \\
\cmidrule(lr){2-3} \cmidrule(lr){4-5}
\cmidrule(lr){6-7} \cmidrule(lr){8-9}
& MSE & MAE & MSE & MAE & MSE & MAE & \cellcolor{gray!20}MSE & \cellcolor{gray!20}MAE\\
\midrule
$1$  & \underline{0.1605} & 0.2652 & 0.1611 & \textbf{0.2548} & \textbf{0.1556} & \underline{0.2641} & \cellcolor{gray!20}0.1294 & \cellcolor{gray!20}0.2592\\
$2$  & \textbf{0.0899} & \textbf{0.1477} & 0.0943 & \underline{0.1639} & \underline{0.0939} & 0.1646 & \cellcolor{gray!20}0.0926 & \cellcolor{gray!20}0.1578\\
$3$  & 0.0106 & \underline{0.0676} & \underline{0.0102} & 0.0692 & \textbf{0.0090} & \textbf{0.0609} & \cellcolor{gray!20}0.0076 & \cellcolor{gray!20}0.0558\\
$4$  & \underline{0.0334} & \underline{0.1282} & 0.0368 & 0.1452 & \textbf{0.0322} & \textbf{0.1248} & \cellcolor{gray!20}0.0319 & \cellcolor{gray!20}0.1227\\
$5$  & 0.2343 & 0.2952 & \underline{0.1951} & \textbf{0.2666} & \textbf{0.1896} & \underline{0.2716} & \cellcolor{gray!20}0.1633 & \cellcolor{gray!20}0.2642\\
$6$  & \underline{0.0236} & \textbf{0.0950} & 0.0251 & 0.1078 & \textbf{0.0233} & \underline{0.0975} & \cellcolor{gray!20}0.0218 & \cellcolor{gray!20}0.0907\\
$7$  & \underline{0.0401} & \underline{0.1323} & 0.0454 & 0.1555 & \textbf{0.0392} & \textbf{0.1311} & \cellcolor{gray!20}0.0400 & \cellcolor{gray!20}0.1304\\
$8$  & \textbf{0.0277} & \textbf{0.1037} & 0.0305 & 0.1152 & \underline{0.0285} & \underline{0.1086} & \cellcolor{gray!20}0.0268 & \cellcolor{gray!20}0.1016\\
$9$  & 0.0777 & 0.1617 & \textbf{0.0400} & \underline{0.1353} & \underline{0.0435} & \textbf{0.1306} & \cellcolor{gray!20}0.0337 & \cellcolor{gray!20}0.1140\\
$10$ & 0.0222 & 0.0943 & \underline{0.0197} & \underline{0.0817} & \textbf{0.0186} & \textbf{0.0761} & \cellcolor{gray!20}0.0187 & \cellcolor{gray!20}0.0778\\
\midrule
\textbf{Overall} 
& 0.0720 & 0.1491 
& \underline{0.0658} & \underline{0.1404} 
& \textbf{0.0633} & \textbf{0.1375} 
& \cellcolor{gray!20}0.0566 & \cellcolor{gray!20}0.1319\\
\bottomrule[1.5pt]
\end{tabular}
\end{table*}

The following methods are compared in the experiments:
\begin{itemize}    
    \item \textbf{FedAvg}: The standard FL algorithm.
    \item \textbf{SLI}: Sequential local initialization strategy without aggregation, where the last local model in the sequence is taken as the next global model.
    \item \textbf{SLIAvg}: Sequential local initialization with aggregation, where all resulting local models are averaged to obtain the next global model.
    \item \textbf{Centralized}: All client datasets are collected and trained centrally. This is reported as a reference baseline and does not satisfy the privacy requirement of FL.
\end{itemize}

Table~\ref{table2} compares the household-level forecasting performance of different training strategies. In this experiment, the number of clients is set to $K=10$, 5 clients participate in each communication round, and the Transformer model is adopted. As presented in Table~\ref{table2}, SLIAvg achieves the best performance for most clients and obtains the lowest overall error, reducing the MSE and MAE from 0.0720 and 0.1491 using FedAvg to 0.0633 and 0.1375, respectively. 
SLI also improves over FedAvg, while SLIAvg further reduces the prediction error through the additional aggregation step.

\begin{table}[H]
\caption{Test Loss Comparison of Different FL Strategies Under Different Numbers of Clients}
\label{table4}
\centering
\renewcommand{\arraystretch}{1.05}
\setlength{\tabcolsep}{5pt}
\begin{tabular}{C{1.4cm} C{1.2cm} C{1.2cm} C{1.2cm} C{1.2cm}}
\toprule[1.5pt]
\multirow{2}{*}{FL Strategy} 
& \multicolumn{2}{c}{K = 15} 
& \multicolumn{2}{c}{K = 30}\\
\cmidrule(lr){2-3} \cmidrule(lr){4-5}
& Transformer & LSTM & Transformer & LSTM \\
\midrule
FedAvg 
& 0.1363 & 0.0897 & 0.1016 & 0.0576\\
SLI 
& 0.0908 & 0.0885 & 0.0749 & 0.0560 \\
SLIAvg 
& \textbf{0.0853} & \textbf{0.0764} & \textbf{0.0744} & \textbf{0.0537} \\
\rowcolor{gray!20}
Centralized 
& 0.0749 & 0.0704 & 0.0573 & 0.0414 \\
\bottomrule[1.5pt]
\end{tabular}
\end{table}

\begin{figure}
\begin{center}
\includegraphics[width=7cm]{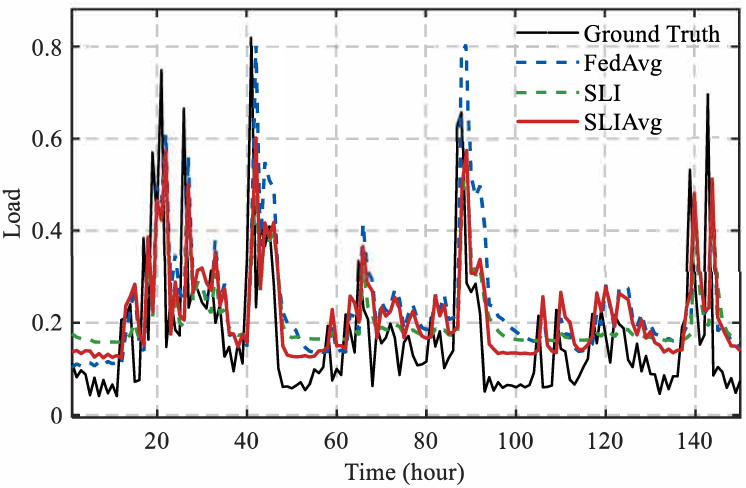}
\end{center}
\caption{STLF result for Client 1.}
\label{fig:load_c1}
\end{figure}

Table~\ref{table4} further compares different federated strategies under different numbers of clients. Compared with FedAvg, SLIAvg substantially narrows the performance gap between federated training and the centralized reference. Figs.~\ref{fig:load_c1} shows representative forecasting trajectories for one client.
These results indicate that SLIAvg helps mitigate the adverse effects of client heterogeneity and improves the prediction accuracy.

\subsection{Privacy Discussion}

\subsubsection{Privacy of Pretrained Initialization}

The auxiliary data used for pretraining initialization are assumed to be privacy-compliant. If load records voluntarily shared by some users are included in the auxiliary dataset, proper anonymization and authorization are required to avoid violating the privacy-preserving setting of FL. Therefore, the proposed pretrained initialization remains compatible with the privacy-preserving requirement of FL. Moreover, it can be combined with existing privacy-enhancing techniques during federated training, since it only modifies the initial global model and does not change the client-side training or server-side aggregation protocol.

\subsubsection{SLIAvg Privacy}
Since raw load data remain local throughout training, the proposed SLIAvg strategy also preserves the privacy-compatible setting of FL. Additional privacy-enhancing techniques can also be incorporated when formal privacy guarantees are required.
Compared with standard parallel FL, SLIAvg introduces extra training latency because the initialization model must be synchronized from one client to the next through the server. 
This requires more training time, especially when many clients participate in each round. 
Nevertheless, this cost does not change the decentralized paradigm of FL, and the trade-off is justified in STLF where forecasting accuracy is prioritized over training speed.

\section{Conclusions}

This paper investigates federated STLF from the joint perspective of client heterogeneity and model initialization. Load data heterogeneity is revealed by analyzing heterogeneous responses to exogenous factors and differences in temporal load profiles. Based on this analysis, two initialization strategies are proposed for federated STLF. First, the pretrained initialization strategy uses auxiliary public load data to initialize the global model before FL. Second, SLIAvg introduces sequential local initialization to promote smoother adaptations across client updates. The proposed strategies are evaluated on real load data using two representative forecasting models, demonstrating their effectiveness in reducing client drift, improving convergence behavior, and achieving better predictive performance.

Future work will focus on reducing the training latency of SLIAvg while preserving its benefits.
\appendices

\bibliographystyle{IEEEtran}
\bibliography{References}
\end{document}